 \documentclass[final,5p,times,twocolumn,authoryear]{elsarticle}

\usepackage{amssymb}
\usepackage{lipsum}
\usepackage{xcolor}%

\begin{document}

\begin{frontmatter}



\title{Large Language Models for Multi-Choice Question Classification of Medical Subjects}


\author[first]{V\'{i}ctor Ponce-L\'{o}pez}
\affiliation[first]{organization={University College London}, Department of Computer Science
            addressline={Centre for Artificial Intelligence, University College London}, 
            city={London},
            postcode={WC1E 6BT}, 
            country={UK}}

\begin{abstract}
The aim of this paper is to evaluate whether large language models trained on multi-choice question data can be used to discriminate between medical subjects. This is an important and challenging task for automatic question answering. To achieve this goal, we train deep neural networks for multi-class classification of questions into the inferred medical subjects. Using our Multi-Question (MQ) Sequence-BERT method, we outperform the state-of-the-art results on the MedMCQA dataset with an accuracy of \textit{0.68} and \textit{0.60} on their development and test sets, respectively. In this sense, we show the capability of AI and LLMs in particular for multi-classification tasks in the Healthcare domain.
\end{abstract}




\end{frontmatter}




\section{Introduction} \label{sec:intro}

One important and challenging area in Natural Language Processing (NLP) is automatic Question Answering (QA) to efficiently access to the vast amount of information available in the format of massive text data. Automatic QA is still a challenge that needs further exploration in real medical examinations given the level of complexity of medical subjects like surgery, pharmacology, or medicine, amongst others. Artificial Intelligence and, in particular, deep neural networks have been key to build appealing large-language models in the in medical-QA to provide a comprehensive understanding of the medical domain~\cite{medmcqa}. 

In this paper, we extend the existing large language models and fine-tune them to evaluate on for automatic QA down-stream task. We apply a current baseline evaluation to classify questions for Medical Multiple-Choice Question Answering for one of the largest datasets in this domain, outperforming the state-of-the-art to predict the outcome of the given questions into the 21 medical subjects.

\section{Materials and Methods}

In this section, we describe the medical multi-choice question answering dataset, methods adopted to fine tune large language models, and the experiments employed for the evaluation of automatic QA.

\subsection{MedMCQA dataset}

The Medical Multiple-Choice Question Answering (MedMCQA)~\cite{medmcqa} is a large-scale dataset designed to address real-world medical entrance exam questions. It contains has More than 194k high-quality AIIMS \& NEET PG entrance exam MCQs covering 2.4k healthcare topics and 21 medical subjects are collected with an average token length of 12.77 and high topical diversity. The dataset contains questions about the following subjects: Anesthesia, Anatomy, Biochemistry, Dental, ENT, Forensic Medicine (FM), Obstetrics and Gynecology, (O\&G), Medicine, Microbiology, Ophthalmology, Orthopedics, Pathology, Pediatrics, Pharmacology, Physiology, Psychiatry, Radiology, Skin, Preventive \& Social Medicine (PSM), Surgery.

External knowledge sources from Wikipedia and Pubmed contributed to provide context. This allowed to understand the contribution of these knowledge sources and the usefulness of the internal knowledge. The dataset is split into 183K train examples, 6K in the development set, and 4K in the test set from the whole set of MCQs. These splits of the dataset are provided for a baseline evaluation. The baseline experiments on this dataset with the current state-of-the-art methods could only answer 47\% of the question correctly, which was far behind the performance of human candidates. The primary motivation of the baseline experiments is to understand the adequacy of the current models in answering multiple-choice questions meant for human domain experts.
 
\subsection{Large Language Model} \label{sec:llm}

LLM embeddings are able to encode the semantics of the text data sequences. They are able to encode contextual and temporal meaningful information into their tensor representations. We use a Sentence-BERT~\cite{sentence-bert} for multi-sequence classification both to extract sequence embeddings from these pretrained language models and to fine-tune them specifically for our downstream MCQA task.

\subsubsection{Multi-Question SequenceBERT} \label{sec:sequence-LLM}

An input to the LLM will be a question $q\in Q$, where $Q$ is a set of questions belonging to multiple subjects. Therefore, the total dimensionality of the feature space in our case will be $N\times D$ for the total number of questions $N$ and the dimensionality $D$ of the LLM embedding tensors. We refer to this LLM approach as Multi-Question (MQ) sequence. 

\textit{Encoding}:
    First, we tokenize each question to encode the input sequences using padding to the longest sequence in the batch, and truncation to the maximum acceptable input length for the model. 
    Then, we compute the output embeddings via mean pooling and apply a functional $L_2$ normalisation by transforming each element 
    of the embedding tensor along its dimension $D$. 
    This results in a set of feature embeddings $E=\{e_1, e_2, ..., e_{N_u}\}$, where each element $e\in E$ is an embedding tensor of length $D$ for the question $q$. 

\subsubsection{Experimental Settings}
    We use the Transformers~\cite{transformers} API. In the tokenizer, we specify the end of sentence token for padding. In a multi-question sequence LLM, we use the pretrained model and encode all the set of questions to extract the embedding tensors. We split this process in batches of $500$ questions each to minimise memory load. The length of the MQ-SequenceBERT' embedding tensors is $384$. We use an standard AdamW~\cite{AdamW} optimiser with no weight decay and a learning rate of $1e-5$ and epsilon $1e-8$ to fine-tune our models for $10$ epochs and $100$ training steps per epoch using a batch size of $8$ per GPU device. We ran our experiments in a NVIDIA A40 GPU.

\section{Results}

\begin{table}
    \begin{tabular}{l c c} 
     \hline
     Method & Dev. & Test \\
     & Accuracy & Accuracy \\ 
     \hline
     BERT~\cite{devlin-etal-2019-bert}$_{BASE}$ & 	0.35 & 0.33 \\
     BioBERT~\cite{biobert-lee} & 	0.38 & 0.37 \\
     SciBERT~\cite{beltagy-etal-2019-scibert} & 	0.39 & 0.39 \\
     PubmedBERT~\cite{Gu2020DomainSpecificLM} & 	0.40 & 0.41 \\
     Codex {\scriptsize{5-shot CoT}}~\cite{Lievin2022CanLL} & 0.63 & \textit{0.60} \\
     \textbf{Our MQ-SequenceBERT} & \textbf{0.68} & \textbf{0.60} \\
     \hline
    \end{tabular}
    \caption{State-of-the-art on multi-class classification accuracy on the development and test sets, showing outperformance of our fine-tuned Multi-Question Sequence-BERT LLM.
    }
    \label{tab:results}
\end{table}

\begin{figure}[t]
	\centering 
	\includegraphics[width=0.5\textwidth]{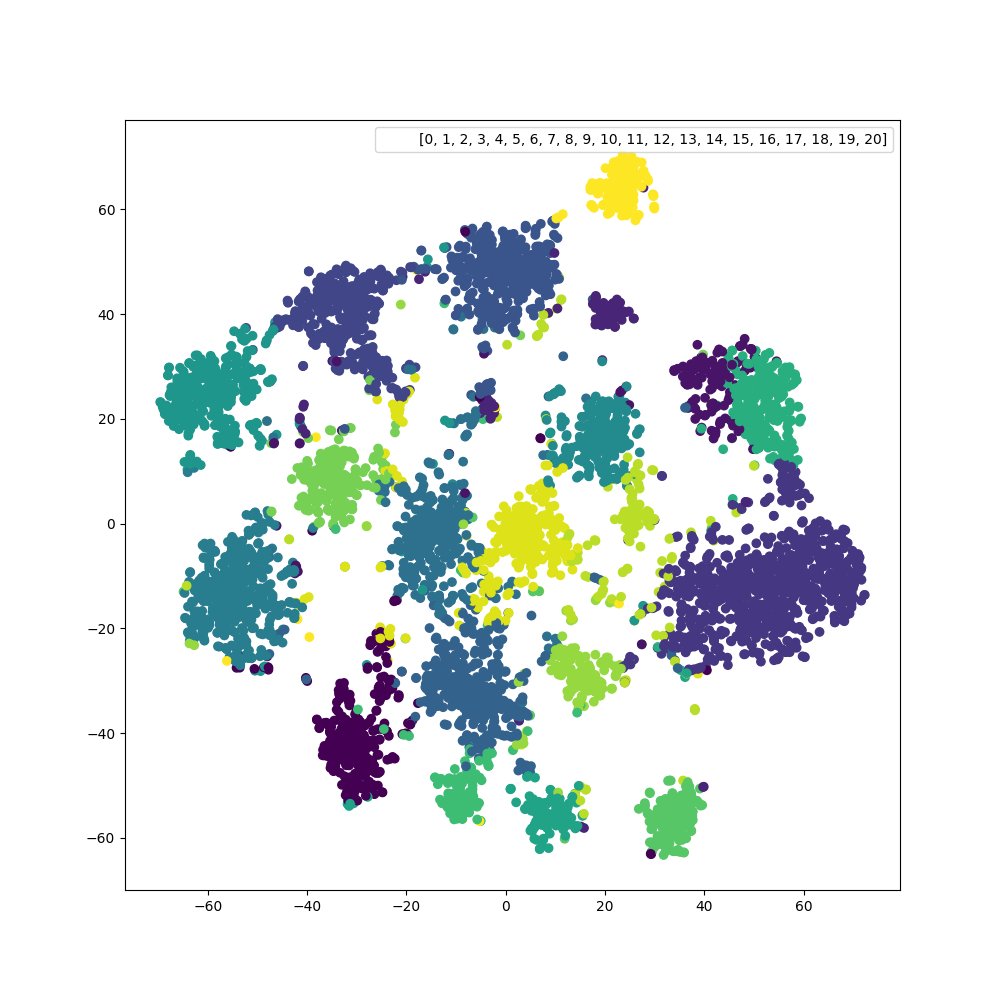}	
	\caption{Distribution of the question groups for the 21 medical subjects represented via 2 components of t-SNE.} 
	\label{fig:embeddings}%
\end{figure}

Table~\ref{tab:results} shows the performance of the state-of-the-art methods for the experiments without using context. It includes our the accuracy both on the development and test sets. Our experiments showed that our MQ-SequenceBERT outperformed the state-of-the-art both clearly on the development set by 5\% and slightly on the sets. We calculate the accuracy from the predictions of the questions into subject labels. To evaluate the test set, we trained a model twice using both the training set and the whole learning set. However, there was no significant difference on the achieve performance for one or the other strategy. Figure~\ref{fig:embeddings} shows a representation of the embeddings using two components of a T-distributed Stochastic Neighbour Embedding (t-SNE) approach. This method allows to show the capability of our method to distinguish amongst the 21 distinct groups belonging to the different medical subjects for the embeddings distribution corresponding to each question. These results show the capability of AI and, in aprticular, fine-tuning large language models, to classify multi-choice questions for automatic QA in the medical domain. 

\section{Conclusion}

We presented a method to fine-tune large language models for automatic question answering in the medical domain. We outperformed the state-of-the-art results on the MedMCQA dataset, one of the most popular datasets for Multi-Choice Question Answering that covers a wide variety of medical subjects and topics. We achieved this results in the baseline evaluation for this dataset without the need of using context information. Future work is aimed to dig into other specific datasets in the medical domain where a comparison of different methods including LLM can be applicable to determine the feasibility of early detection systems to support clinical decision-making systems as well as experts in the clinical domain.

\section*{Acknowledgements}

This work has been possible and supported by i-sense, the Engineering and Physical Sciences Research Council (EPSRC) International Research Council (IRC) in Early Warning Sensing Systems for Infectious Diseases.

\bibliographystyle{elsarticle-harv} 
\bibliography{references}

\begin{thebibliography}{9}
\expandafter\ifx\csname natexlab\endcsname\relax\def\natexlab#1{#1}\fi
\providecommand{\url}[1]{\texttt{#1}}
\providecommand{\href}[2]{#2}
\providecommand{\path}[1]{#1}
\providecommand{\DOIprefix}{doi:}
\providecommand{\ArXivprefix}{arXiv:}
\providecommand{\URLprefix}{URL: }
\providecommand{\Pubmedprefix}{pmid:}
\providecommand{\doi}[1]{\href{http://dx.doi.org/#1}{\path{#1}}}
\providecommand{\Pubmed}[1]{\href{pmid:#1}{\path{#1}}}
\providecommand{\bibinfo}[2]{#2}
\ifx\xfnm\relax \def\xfnm[#1]{\unskip,\space#1}\fi
\bibitem[{Beltagy et~al.(2019)Beltagy, Lo and Cohan}]{beltagy-etal-2019-scibert}
\bibinfo{author}{Beltagy, I.}, \bibinfo{author}{Lo, K.}, \bibinfo{author}{Cohan, A.}, \bibinfo{year}{2019}.
\newblock \bibinfo{title}{{S}ci{BERT}: A pretrained language model for scientific text}, in: \bibinfo{editor}{Inui, K.}, \bibinfo{editor}{Jiang, J.}, \bibinfo{editor}{Ng, V.}, \bibinfo{editor}{Wan, X.} (Eds.), \bibinfo{booktitle}{Proceedings of the 2019 Conference on Empirical Methods in Natural Language Processing and the 9th International Joint Conference on Natural Language Processing (EMNLP-IJCNLP)}, \bibinfo{publisher}{Association for Computational Linguistics}, \bibinfo{address}{Hong Kong, China}. pp. \bibinfo{pages}{3615--3620}.
\newblock \URLprefix \url{https://aclanthology.org/D19-1371}, \DOIprefix\doi{10.18653/v1/D19-1371}.
\bibitem[{Devlin et~al.(2019)Devlin, Chang, Lee and Toutanova}]{devlin-etal-2019-bert}
\bibinfo{author}{Devlin, J.}, \bibinfo{author}{Chang, M.W.}, \bibinfo{author}{Lee, K.}, \bibinfo{author}{Toutanova, K.}, \bibinfo{year}{2019}.
\newblock \bibinfo{title}{{BERT}: Pre-training of deep bidirectional transformers for language understanding}, in: \bibinfo{editor}{Burstein, J.}, \bibinfo{editor}{Doran, C.}, \bibinfo{editor}{Solorio, T.} (Eds.), \bibinfo{booktitle}{Proceedings of the 2019 Conference of the North {A}merican Chapter of the Association for Computational Linguistics: Human Language Technologies, Volume 1 (Long and Short Papers)}, \bibinfo{publisher}{Association for Computational Linguistics}, \bibinfo{address}{Minneapolis, Minnesota}. pp. \bibinfo{pages}{4171--4186}.
\newblock \URLprefix \url{https://aclanthology.org/N19-1423}, \DOIprefix\doi{10.18653/v1/N19-1423}.
\bibitem[{Gu et~al.(2020)Gu, Tinn, Cheng, Lucas, Usuyama, Liu, Naumann, Gao and Poon}]{Gu2020DomainSpecificLM}
\bibinfo{author}{Gu, Y.}, \bibinfo{author}{Tinn, R.}, \bibinfo{author}{Cheng, H.}, \bibinfo{author}{Lucas, M.R.}, \bibinfo{author}{Usuyama, N.}, \bibinfo{author}{Liu, X.}, \bibinfo{author}{Naumann, T.}, \bibinfo{author}{Gao, J.}, \bibinfo{author}{Poon, H.}, \bibinfo{year}{2020}.
\newblock \bibinfo{title}{Domain-specific language model pretraining for biomedical natural language processing}.
\newblock \bibinfo{journal}{ACM Transactions on Computing for Healthcare (HEALTH)} \bibinfo{volume}{3}, \bibinfo{pages}{1 -- 23}.
\newblock \URLprefix \url{https://api.semanticscholar.org/CorpusID:220919723}.
\bibitem[{Lee et~al.(2019)Lee, Yoon, Kim, Kim, Kim, So and Kang}]{biobert-lee}
\bibinfo{author}{Lee, J.}, \bibinfo{author}{Yoon, W.}, \bibinfo{author}{Kim, S.}, \bibinfo{author}{Kim, D.}, \bibinfo{author}{Kim, S.}, \bibinfo{author}{So, C.H.}, \bibinfo{author}{Kang, J.}, \bibinfo{year}{2019}.
\newblock \bibinfo{title}{{BioBERT: a pre-trained biomedical language representation model for biomedical text mining}}.
\newblock \bibinfo{journal}{Bioinformatics} \bibinfo{volume}{36}, \bibinfo{pages}{1234--1240}.
\newblock \URLprefix \url{https://doi.org/10.1093/bioinformatics/btz682}, \DOIprefix\doi{10.1093/bioinformatics/btz682}.
\bibitem[{Li'evin et~al.(2022)Li'evin, Hother and Winther}]{Lievin2022CanLL}
\bibinfo{author}{Li'evin, V.}, \bibinfo{author}{Hother, C.E.}, \bibinfo{author}{Winther, O.}, \bibinfo{year}{2022}.
\newblock \bibinfo{title}{Can large language models reason about medical questions?}
\newblock \bibinfo{journal}{ArXiv} \bibinfo{volume}{abs/2207.08143}.
\newblock \URLprefix \url{https://api.semanticscholar.org/CorpusID:250627547}.
\bibitem[{Loshchilov and Hutter(2017)}]{AdamW}
\bibinfo{author}{Loshchilov, I.}, \bibinfo{author}{Hutter, F.}, \bibinfo{year}{2017}.
\newblock \bibinfo{title}{Decoupled weight decay regularization}, in: \bibinfo{booktitle}{International Conference on Learning Representations}.
\newblock \URLprefix \url{https://api.semanticscholar.org/CorpusID:53592270}.
\bibitem[{Pal et~al.(2022)Pal, Umapathi and Sankarasubbu}]{medmcqa}
\bibinfo{author}{Pal, A.}, \bibinfo{author}{Umapathi, L.K.}, \bibinfo{author}{Sankarasubbu, M.}, \bibinfo{year}{2022}.
\newblock \bibinfo{title}{Medmcqa: A large-scale multi-subject multi-choice dataset for medical domain question answering}, in: \bibinfo{editor}{Flores, G.}, \bibinfo{editor}{Chen, G.H.}, \bibinfo{editor}{Pollard, T.}, \bibinfo{editor}{Ho, J.C.}, \bibinfo{editor}{Naumann, T.} (Eds.), \bibinfo{booktitle}{Proceedings of the Conference on Health, Inference, and Learning}, \bibinfo{publisher}{PMLR}. pp. \bibinfo{pages}{248--260}.
\newblock \URLprefix \url{https://proceedings.mlr.press/v174/pal22a.html}.
\bibitem[{Reimers and Gurevych(2019)}]{sentence-bert}
\bibinfo{author}{Reimers, N.}, \bibinfo{author}{Gurevych, I.}, \bibinfo{year}{2019}.
\newblock \bibinfo{title}{Sentence-{BERT}: Sentence {E}mbeddings using {S}iamese {BERT}-networks}, in: \bibinfo{booktitle}{Proceedings of the 2019 Conference on Empirical Methods in Natural Language Processing}, \bibinfo{publisher}{Association for Computational Linguistics}, \bibinfo{address}{Online}.
\newblock \URLprefix \url{https://arxiv.org/abs/1908.10084}.
\bibitem[{Wolf et~al.(2020)Wolf, Debut, Sanh, Chaumond, Delangue, Moi, Cistac, Rault, Louf, Funtowicz, Davison, Shleifer, von Platen, Ma, Jernite, Plu, Xu, Scao, Gugger, Drame, Lhoest and Rush}]{transformers}
\bibinfo{author}{Wolf, T.}, \bibinfo{author}{Debut, L.}, \bibinfo{author}{Sanh, V.}, \bibinfo{author}{Chaumond, J.}, \bibinfo{author}{Delangue, C.}, \bibinfo{author}{Moi, A.}, \bibinfo{author}{Cistac, P.}, \bibinfo{author}{Rault, T.}, \bibinfo{author}{Louf, R.}, \bibinfo{author}{Funtowicz, M.}, \bibinfo{author}{Davison, J.}, \bibinfo{author}{Shleifer, S.}, \bibinfo{author}{von Platen, P.}, \bibinfo{author}{Ma, C.}, \bibinfo{author}{Jernite, Y.}, \bibinfo{author}{Plu, J.}, \bibinfo{author}{Xu, C.}, \bibinfo{author}{Scao, T.L.}, \bibinfo{author}{Gugger, S.}, \bibinfo{author}{Drame, M.}, \bibinfo{author}{Lhoest, Q.}, \bibinfo{author}{Rush, A.M.}, \bibinfo{year}{2020}.
\newblock \bibinfo{title}{Transformers: State-of-the-art natural language processing}, in: \bibinfo{booktitle}{Proceedings of the 2020 Conference on Empirical Methods in Natural Language Processing: System Demonstrations}, \bibinfo{publisher}{Association for Computational Linguistics}, \bibinfo{address}{Online}. pp. \bibinfo{pages}{38--45}.
\newblock \URLprefix \url{https://www.aclweb.org/anthology/2020.emnlp-demos.6}.

\end{thebibliography}






\end{document}